\documentclass{article}
\usepackage{caption}

\PassOptionsToPackage{numbers, compress}{natbib}

\usepackage{graphicx}
\usepackage{wrapfig}   

\usepackage[utf8]{inputenc} 
\usepackage[T1]{fontenc}    
\usepackage{url}            
\usepackage{booktabs}       
\usepackage{amsfonts}       
\usepackage{nicefrac}       
\usepackage{microtype}      

\usepackage{url}

\usepackage{microtype}
\usepackage{graphicx}
\usepackage{subfigure}
\usepackage{makecell}
\usepackage{booktabs} 
\usepackage{hyperref}
\usepackage[table]{xcolor}
\definecolor{aliceblue}{RGB}{176,223,229}

\usepackage{animate}
\usepackage{listings}
\usepackage{algorithm}
\usepackage{algpseudocode}
\usepackage{enumitem}
\usepackage{silence}
\usepackage{booktabs}
\usepackage{multirow}

\usepackage{xcolor}         

\usepackage{amsmath}
\usepackage{amssymb}

\usepackage[final]{neurips_2024}

\usepackage[utf8]{inputenc} 
\usepackage[T1]{fontenc}    
\usepackage{hyperref}       
\usepackage{url}            
\usepackage{booktabs}       
\usepackage{amsfonts}       
\usepackage{nicefrac}       
\usepackage{microtype}      
\usepackage{xcolor}

\def\model_name{MagicInfinite}  

\title{MagicInfinite: Generating Infinite Talking Videos with Your Words and Voice}

\author{	
Hongwei Yi$^{1}$\footnotemark[1] \ \footnotemark[2] ~~~ Tian Ye$^{1,3}$\footnotemark[1] ~~~ Shitong Shao$^{1,3}$\footnotemark[1] ~~~ Xuancheng Yang$^{1}$\footnotemark[1] ~~~ Jiantong Zhao$^{1}$\footnotemark[1] \\ \\
\textbf{Hanzhong Guo}$^{1,4}$\footnotemark[1] ~~~ \textbf{Terrance Wang}$^{1}$\footnotemark[1] ~~~ \textbf{Qingyu Yin}$^{1}$~~~ \textbf{Zeke Xie}$^{3}$ ~~~ \textbf{Lei Zhu}$^{3}$ \\ \\ \textbf{Wei Li}$^{1}$ ~~~
\textbf{Michael Lingelbach}$^{1}$ ~~~\textbf{Daquan Zhou}$^{2}$\footnotemark[3] \\ \\
$^1$ Hedra Inc.  \quad $^2$ Peking University \quad $^3$ HKUST(GZ) \quad $^4$ HKU \\
}

\begin{document}

\maketitle

\begin{abstract}
We present {\model_name}, a novel diffusion Transformer (DiT) framework that overcomes traditional portrait animation limitations, delivering high-fidelity results across diverse character types—realistic humans, full-body figures, and stylized anime characters. It supports varied facial poses, including back-facing views, and animates single or multiple characters with input masks for precise speaker designation in multi-character scenes.
Our approach tackles key challenges with three innovations: (1) 3D full-attention mechanisms with a sliding window denoising strategy, enabling infinite video generation with temporal coherence and visual quality across diverse character styles; (2) a two-stage curriculum learning scheme, integrating audio for lip sync, text for expressive dynamics, and reference images for identity preservation, enabling flexible multi-modal control over long sequences; and (3) region-specific masks with adaptive loss functions to balance global textual control and local audio guidance, supporting speaker-specific animations. Efficiency is enhanced via our innovative unified step and cfg distillation techniques, achieving a 20× inference speed boost over the basemodel—generating a 10-second 540x540p video in 10 seconds or 720x720p in 30 seconds on 8 H100 GPUs—without quality loss.
Evaluations on our new benchmark demonstrate {\model_name}’s superiority in audio-lip synchronization, identity preservation, and motion naturalness across diverse scenarios. It is publicly available at \url{https://www.hedra.com/}, with examples at \url{https://magicinfinite.github.io/}.

\end{abstract}


\begin{figure}[t]
    \centering
    \includegraphics[width=1.0 \textwidth]{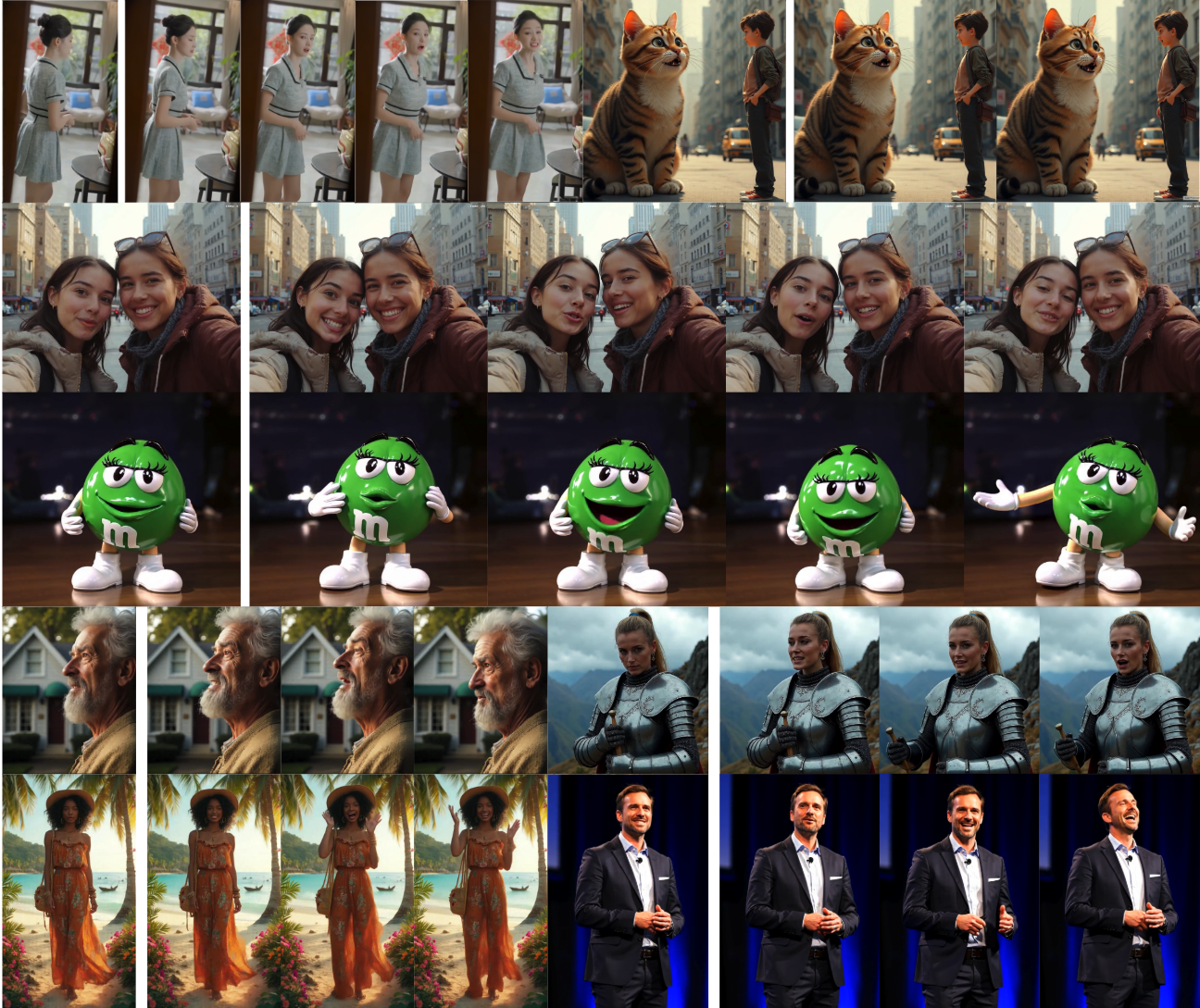}
    \caption{
        Given a portrait image, our model can generate compelling, realistic, and vivid animation videos with control over text and voice, ensuring temporal coherence and perceptual quality even under significant head pose variations and diverse portrait styles.
    }
    \label{fig:teaser}
\end{figure}

\renewcommand{\thefootnote}{}
\footnotetext{$^{*}$Equal contribution, $^{\dagger}$Project lead, $^\ddagger$Corresponding author.}
\renewcommand{\thefootnote}{\arabic{footnote}}

\section{INTRODUCTION}
Talking avatars—realistic digital characters created from a reference image using audio or text input—are a transformative technology at the intersection of computer vision and human-computer interaction, driving advancements in digital entertainment, education, and AI communication~\cite{ma2021pixel, wang2021one, kal2024educational, rehm2016role, aicommunicate, johnson2018assessing}, fundamentally reshaping how humans interact with digital content.

Pioneering works~\cite{su2024audio, song2022audio, ye2024real3d, Chatziagapi2023, zhang2023sadtalker, ma2023styletalk, xing2023codetalker, bai2024efficient, peng2023emotalk, cho2024gaussiantalker, chen2024gstalker, li2024talkinggaussian, zhuang2024learn2talk, peng2024synctalk, li2024ae, liu2023moda, guan2023stylesync, wang2023seeing} use neural networks and rendering like GANs~\cite{goodfellow2020generative}, NeRF~\cite{mildenhall2021nerf}, and Gaussian Splatting~\cite{kerbl20233d} to fuse motion and identity features for high-fidelity talking heads. Yet, 3DMM~\cite{tran2018_3DMM} and FLAME~\cite{li2017_FLAME} struggle with motion accuracy, and rendering limits resolution and quality.
Recent LDMs~\cite{rombach2022LDM} boost video synthesis~\cite{blattmann2023SVD, guo2023animatediff, chen2023pixart, xu2024easyanimate, wang2024magicvideo, yang2024cogvideox, zhou2024allegro, polyak2024movieGen, kong2024hunyuanvideo} with better diversity and coherence. We apply T2V diffusion to portrait animation. Studies~\cite{tian2024emo, wang2024V-Express, chen2024echomimic, yang2024megactor, jiang2024loopy, zheng2024memo} use LDM priors~\cite{rombach2022LDM} to model spatial and temporal dimensions separately, but falter in coherence for large movements, non-frontal faces, or high-resolution cases due to weak frame correspondence~\cite{xu2024easyanimate}.

In this work, we propose \textbf{{\model_name}}, a novel zero-shot framework that utilizes 3D full-attention to effectively model video along both temporal and spatial dimensions. \textbf{{\model_name}} demonstrates superior perceptual quality and temporal coherence across various situational motions and actions. 
We explored the simultaneous use of text and voice for synthesizing portrait animation videos, where the voice primarily provides lip movement dynamics, while the text supplies the speaker's expressions, actions, and background transformations.
However, the integration of both audio and textual prompts as motion signals presents certain challenges: 
The pre-trained T2V model leverages 3D full-attention mechanisms, attending to all tokens from both the video modality and textual prompts, thereby capturing a robust dependency between the entire video content and the associated text.
The text plays a significant role in controlling the video content, as each video token is related to the textual tokens. 
By contrast, the control of character motion by audio primarily focuses on local video regions, such as lip movements, which are only associated with a small subset of video tokens. 
This creates a conflict between the driving audio and the textual prompt controls.
Therefore, when building upon the traditional form of incorporating audio in prior portrait animation works~\cite{tian2024emo}, which applies cross-attention~\cite{vaswani2017attention_is_all} between audio features and vision representations within a spatial latent space directly, without further manipulation, the model tends to overlook the driving audio's influence on lip movements.
This limitation becomes even more apparent when the face occupies a smaller portion of the image.

To address this challenge, we design a two-stage curriculum learning scheme to effectively enable audio control, guiding the model to progressively learn both textual and audio control. 
Specifically, in the first training phase, only the portrait image and textual prompt are introduced, allowing the model to learn text-controlled image-to-video (I2V) generation. 
In the second phase, both text and audio are incorporated. Here, we utilize a face region mask and an adaptive loss function to guide the conditional cross-attention of audio, specifically targeting local facial movements around the mouth. 
This training scheme effectively integrates the control provided by textual prompts and driving audio for portrait animation.

Owing to the substantial increase in the number of function evaluations (NFEs) caused by multi-step sampling and classifier-free guidance (CFG)~\cite{ho2022cfg, chung2024cfg++}, diffusion models often suffer from sluggish inference speeds—an issue that has been largely neglected in earlier studies. Previous work~\cite{guo2024real_hanzhong} obtained an avatar-based few-step generator by employing latent consistency model (LCM)~\cite{luo2023LCM}, which effectively condensed multiple sampling steps into a reduced number, thereby minimizing the sampling process. However, we found that utilizing LCM for {\model_name} step distillation inevitably led to noticeable degradation in visual quality, often manifesting as significant blurring. To address this pressing concern, we turn to DMD2~\cite{yin2024DMD2}, which has been rigorously validated across numerous image diffusion models~\citep{SDXL} and has consistently demonstrated superior performance compared to LCM, to enable accelerated sampling. Our approach achieves a groundbreaking advancement by simultaneously eliminating \textbf{{\model_name}}'s reliance on CFG~\citep{meng2023distillation} during the step distillation process in DMD2, directly reducing NFEs from 50 to just 4. To circumvent the GPU memory constraints posed by importing three models simultaneously in vanilla DMD2, we adopt LoRA to update the parameters of the fake data distribution estimator, thereby ensuring an efficient and resource-aware training process.

We conducted a thorough evaluation of \textbf{{\model_name}} using a rigorous benchmark we developed, called the \textbf{{\model_name}}-Benchmark. This benchmark consists of 25 driving audio clips spanning diverse speaking scenarios—such as singing, speeches, and rapping—paired with 20 textual prompts describing the speaker’s emotions, actions, and background variations. Additionally, the \textbf{{\model_name}}-Benchmark includes a varied collection of portrait images featuring different styles, face orientations relative to the camera, and face sizes. Under this benchmark, \textbf{{\model_name}} exhibits robust performance across a wide range of portrait images and motion scenarios, effectively driven by both audio and textual prompts, while maintaining efficient inference speeds.

The key contributions of our work are as follows:

\begin{itemize}
\item A novel zero-shot talking avatar framework using 3D full-attention in a pre-trained video DiT with a sliding window denoising strategy, enabling infinite video generation with strong temporal coherence and visual quality across diverse portrait styles and speaking scenarios.

\item A two-stage curriculum learning approach, incorporating face region-guided cross-attention and adaptive loss weighting, which effectively balances global text-based control with local audio-driven control, enabling high-fidelity joint conditioning from both modalities.

\item An innovative strategy for synergistic step and classifier-free guidance (CFG) distillation, combined with a sliding window mechanism, allowing seamless sampling of arbitrarily long videos. This advancement yields a 20$\times$ increase in inference speed while preserving competitive quality metrics, overcoming a key limitation in large-scale diffusion models for portrait animation.
\end{itemize}


\begin{figure*}
	\centering
	\includegraphics[width=0.99\linewidth]{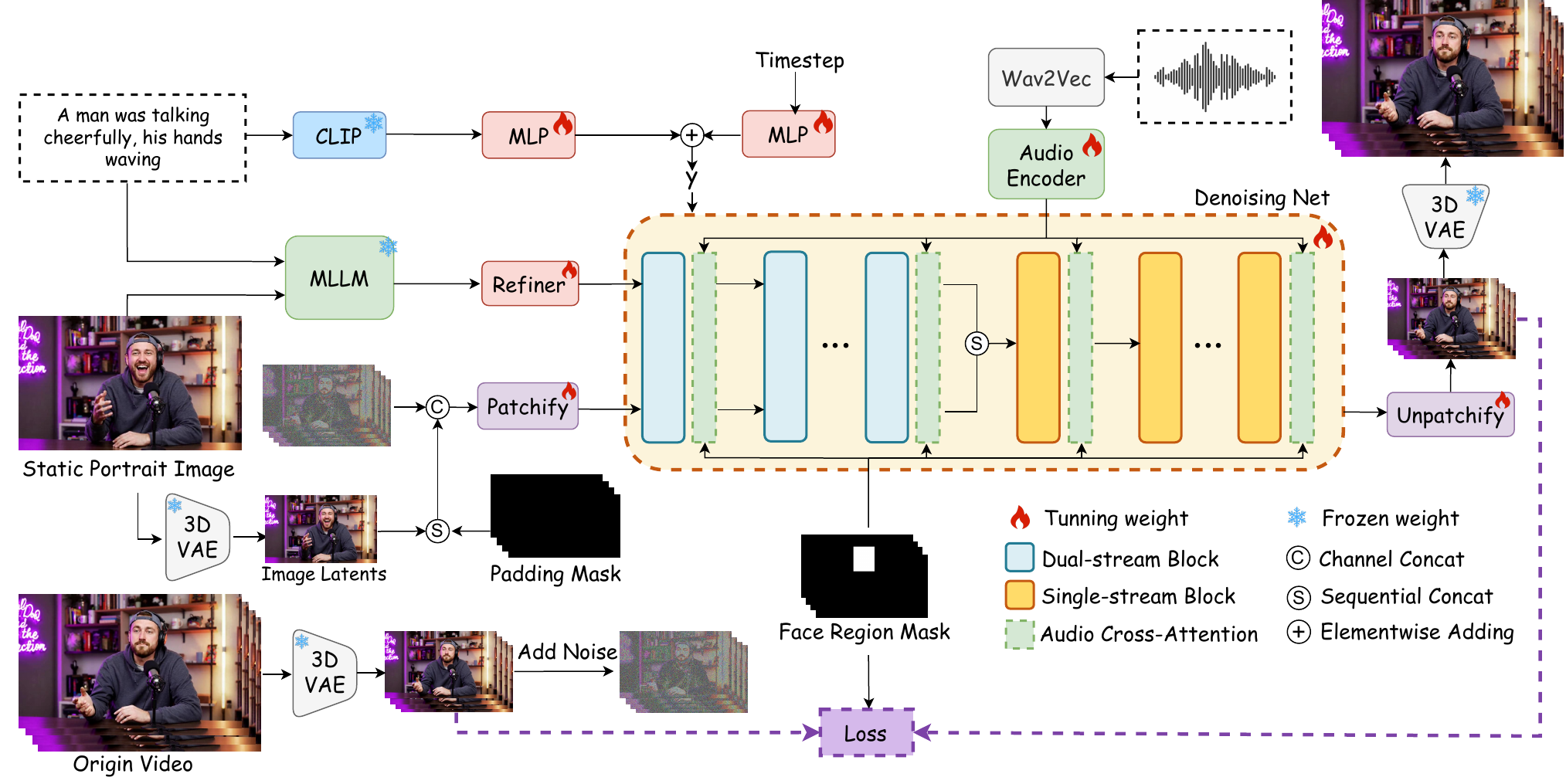}
	\centering
	\caption{
    \textbf{Overview of {\model_name}.} {\model_name} employs a hybrid dual-to-single-stream denoising network with Audio Cross-Attention in final blocks. MLLM encodes static portrait and text into tokens, concatenated for T2V, refined, and denoised. Wav2Vec encodes audio, resampled by an Audio Encoder, and guided by a Face Region Mask for precise lip sync and adaptive loss.
    } 
    \label{fig:method}
\end{figure*}

\section{RELATED WORK}
\subsection{GAN-based Portrait Animation}
The goal of portrait animation is to generate talking videos driven by motion signals. Traditional methods typically employ neural networks to extract motion features from motion signals. These motion features are then transformed into intermediate representations such as landmarks~\cite{su2024audio, song2022audio}, 3D head parameters~\cite{ye2024real3d, Chatziagapi2023, zhang2023sadtalker, ma2023styletalk, xing2023codetalker, bai2024efficient} (3DMM~\cite{tran2018_3DMM}), faces learned with an articulated model and expressions~\cite{peng2023emotalk} (FLAME~\cite{li2017_FLAME}, 3D Gaussian parameters~\cite{cho2024gaussiantalker, chen2024gstalker, li2024talkinggaussian, zhuang2024learn2talk}, 3D tri-plane hash representations~\cite{li2023efficient}, or latent representations~\cite{peng2024synctalk, li2024ae, liu2023moda, guan2023stylesync, wang2023seeing}. Rendering techniques, such as GANs~\cite{goodfellow2020generative}, NeRF~\cite{mildenhall2021nerf}, and Gaussian splatting~\cite{kerbl20233d} are employed to project these intermediate representations into dynamic portrait animations. Despite notable successes, limitations in rendering methods and feature extractors hinder the generation of realistic portrait animation videos.
\subsection{Diffusion-based Portrait Animation}
Latent Diffusion Models~\cite{rombach2022LDM} (LDMs) have demonstrated strong capabilities in image and video generation~\cite{blattmann2023SVD, guo2023animatediff, chen2023pixart, xu2024easyanimate, wang2024magicvideo}. Models such as EMO~\cite{tian2024emo} and V-Express~\cite{wang2024V-Express} leverage audio cues for lip synchronization and weak visual signals for head motion, enabling the generation of natural portrait animation videos. Echomimic~\cite{chen2024echomimic} and MegActor-Sigma~\cite{yang2024megactor} combine both audio and video to offer finer control over the animation process. Loopy~\cite{jiang2024loopy} and MEMO~\cite{zheng2024memo} enhance temporal consistency and emotional expressiveness by integrating modules for audio-to-emotion conversion and past-frame fusion during long video generation. However, these methods face limitations in video fidelity and narrative coherence due to separate attention computations in spatial and temporal domains. 
DiT-based diffusion methods, such as OminiHuman-1~\cite{ominihuman}, CogVideoX~\cite{yang2024cogvideox}, Allegro~\cite{zhou2024allegro} Movie Gen~\cite{polyak2024movieGen}, and HunyuanVideo~\cite{kong2024hunyuanvideo}, employ 3D Full-Attention mechanisms, yielding high-quality video generation. 
Inspired by previous works, we applied the 3D Full-Attention architecture to portrait animation video generation and achieved remarkable results. Our concurrent work, Hallo3~\cite{cui2024hallo3}, was outperformed by {\model_name} in terms of video length and resolution.

\subsection{Efficiency Inference}
Diffusion model inference is inherently limited by its multi-step sampling process, which impacts computational efficiency. To address this challenge, step distillation techniques have emerged as essential approaches for accelerating inference by reducing the number of function evaluations (NFEs). The current research landscape in this domain can be categorized into two primary strategies: aligning few steps with many steps and distribution matching distillation.

In the first category, LCM~\citep{luo2023LCM} has established itself as a prominent framework that effectively aligns few-step models with their multi-step counterparts. This approach has been widely adopted, leading to the development of several open-source few-step VDMs, including MCM~\citep{mcm_accelerate}, the T2V-turbo series~\citep{t2vturbov2}, and FastVideo~\citep{fastvideo}.

Meanwhile, the distribution matching paradigm has demonstrated superior performance in image synthesis compared to LCM-based methods, as evidenced by techniques such as DMD~\citep{dmd}, DMD2~\citep{yin2024DMD2}, and SiD-LSG~\citep{sid_lsg}. Despite its promising results in image generation, the application of distribution matching to video synthesis remains relatively underexplored, with limited implementations such as~\citep{yin2024DMD2}.

\section{METHOD}
Given a single static portrait, our objective is to generate a head animation video that is conditioned on both driving audio and a textual prompt. 
The resulting video should preserve the identity of the subject and the background content depicted in the static portrait while accurately reflecting lip movements, as well as head and facial gestures, and background variations in the portrait image, as dictated by the driving audio and textual prompt. 
In this work, we leverage our internal powerful generative prior of the production-ready DiT-based video generative model, i.e., the text-to-video (T2V) model, integrating control from the static portrait, audio, and textual prompt~\ref{sub_preliminaries}. 
We introduce a cross-attention layer into the model to enable the injection of driving audio~\ref{sub_model_architecture}. 
Building upon this, we propose a novel stepwise training strategy to enable the model to learn control from both voice and text, using the face region mask and adaptive loss function to guide the conditional cross-attention for audio~\ref{sub_curriculum_learn}. 
Finally, we detail our video model distillation techniques, and inference strategies for efficient long video inference~\ref{sub_model_acceleration}.
\subsection{Preliminaries}
\label{sub_preliminaries}
\subsubsection{Flow Matching Model}
Flow Matching~\cite{lipman2022flow_match} transforms a complex probability distribution into a simpler one via the probability density function, enabling the generation of new data samples through inverse transformations. Stable Diffusion V3~\cite{esser2024sd3} is a subclass of Flow Matching models that operate in the latent space, leveraging a pre-trained AutoEncoder~\cite{kingma2013VAE} to facilitate this process.
Unlike standard Text-to-Image (T2I) models, which are conditioned solely on textual inputs $T_s$, {\model_name} derives scene context and motion constraints from the textual prompt $T_s$, the driving audio $A$ and a static portrait image $I_s$. The model trained to learn the reverse transformations of Flow Matching with the objective,
\begin{equation}
\label{loss_func}
    L_{base} =\mathbb{E}_{z_0,z_1,t \sim [0,1]} \bigg[ \Big\lVert {v_t}-{u_\theta} \big(z_t,t,T_s,I_s,A\big) \Big\lVert_2^2 \bigg],
\end{equation} 
where $u_\theta$ is a trainable denoising net. $z_1$ and $z_0$ notes the latent embedding of the training sample and the initialized noise sample drawn from the Gaussian distribution $\mathcal{N}(0,1)$. $z_t$ is the training sample constructed using a linear interpolation. $\mathbf{v}_t=d\mathbf{z}_t/dt=z_1-z_0$ is the velocity which is the target of the model prediction.

\subsubsection{T2V model}
The naive T2V model consists of a Causal 3D VAE~\cite{kingma2013VAE} and a diffusion backbone (denoising net). The 3D VAE compresses pixel-space videos and images into a compact latent space, reducing the token count for the subsequent diffusion transformer model. The diffusion backbone follows a "dual-stream to single-stream" hybrid model design, similar to ~\cite{esser2024sd3}. In the dual-stream phase, video and text tokens are processed independently, and multimodal information is fused in the single-stream phase. Both phases employ a unified Full-Attention mechanism, which models both spatial and temporal dimensions, ensuring strong alignment between text and video content, and guaranteeing high visual quality, motion dynamics, and text-video alignment.

\subsection{Model Architecture}
\label{sub_model_architecture}
To achieve control over the static portrait, we extend the T2V model to an image-to-video (I2V) model, as shown in Fig. ~\ref{fig:method}. Specifically, we treat the first frame of a video as a static portrait image and apply zero-padding to create a tensor of the same shape as the latent input. We utilize a binary mask to encode temporal position information and adjust the parameters of the first convolutional module by zero-initialization.

To further preserve the identity of the subject and the background content depicted in the static portrait image, we use a pre-trained Multimodal Large Language Model (MLLM)~\cite{liu2024llava} to encode the static portrait image. The encoded representations of both the static portrait image and the textual prompt are concatenated along the sequence dimension and used as text tokens.

The clip of driving audio is passed through the Wav2Vec~\cite{baevski2020wav2vec} feature extraction module to obtain audio features. The Audio Encoder then processes these features for resampling, producing latent audio features, which are injected into the denoising net via cross-attention. We selectively insert cross-attention layers at the end of both the single and double blocks, denoted as audio cross-attention, for the controlling of driving audio. The video tokens undergo cross-attention with the latent audio features, performed independently between frames in the latent space.
\subsection{Curriculum Learning Scheme}
\label{sub_curriculum_learn}
Training with both textual prompts and audio often leads to the model neglecting audio control, resulting in incorrect lip synchronization. This occurs because the pre-trained T2V model has already established a strong association between video and text tokens. However, the driving audio primarily focuses on fine details of lip movement, which occupies only a small portion of the video frame. During fine-tuning on new data with supervision via MSE loss, the model tends to further strengthen the association between text and video, particularly focusing on head movements and background changes, while neglecting the alignment of small lip regions. Thus, the key to effectively achieving audio control over lip movements is to enhance the influence of driving audio. To address this, we design a novel two-stage curriculum learning scheme.

In the first stage, we feed driving image and textual prompts into the model through channel-wise concatenation with noise latents and Full-Attention. At this stage, the denoising net remains the same as the T2V model, without audio cross-attention blocks. The training objective is to predict target head animation video in the latent space of 3D VAE, guided by the static portrait image and textual prompt.

In the second stage, we introduce driving audio into the model. Define $M_{face} \in \mathbb{R}^{H{\times}W}$ as the face region mask (the union of facial movement regions across all frames), a binary mask, where $H$, and $W$ represent the height and width of the compressed video frame. 
To emphasize the control of the driving audio, face region mask is applied to the output of the audio cross-attention layer at the pixel level, which can be described as:
\begin{equation}
    {h_i}={h_i}+Attn_{audio}(h_i, A_e, A_e) \times M_{face}'
\end{equation}
$h_i$ is the hidden states of the video after $i$-th single or double block in denoising net. $A_e$ is the feature representation encoded by the audio encoder. $M_{face}'$represents $M_{face}$ reformatted into tokens, specifying which tokens within the ensemble of video modalities are to be aligned with the corresponding $A_e$. $*$ stands for element-wise multiplication.
This forces the audio cross-attention layer to focus on the correlation between driving audio and local facial movements, especially lip movements. Additionally, we design an adaptive loss function that amplifies the loss weight in the facial region, based on the size of the face area. Denote the target latents are of shape $z_1 \in \mathbb{R}^{T{\times}C{\times}H{\times}W}$, where $T$, $C$, $H$, and $W$ represent the frame, channel, height, and width of the compressed video, respectively. The adaptive loss can be written as:
\begin{equation}
\label{loss_func_adap}
    L_{adap} = L_{base}  \frac{H{\times}W}{\sum_{i=0}^{H-1} \sum_{j=0}^{W-1} M_{face}^{ij}},
\end{equation}.
Thus, the total loss function can be written as:
\begin{equation}
\label{loss_func_total}
    L_{total} = L_{base}+\lambda_\textrm{adap}L_{adap}
\end{equation}
while $\lambda_\textrm{adap}$ represents the weight of the adaptive loss.

This ensures that when the facial region is smaller, the loss weight for the facial area is higher, causing the model to focus more on the correlation between driving audio and local facial movements. Conversely, when the facial region is larger, the model focuses more on the correlation between audio and the overall motion. This approach allows the model to better learn the fine-grained control of audio over lip movements.

\subsection{Model Acceleration}
\label{sub_model_acceleration}

\begin{figure}[!h]
    \centering
    \includegraphics[width=1.0\linewidth]{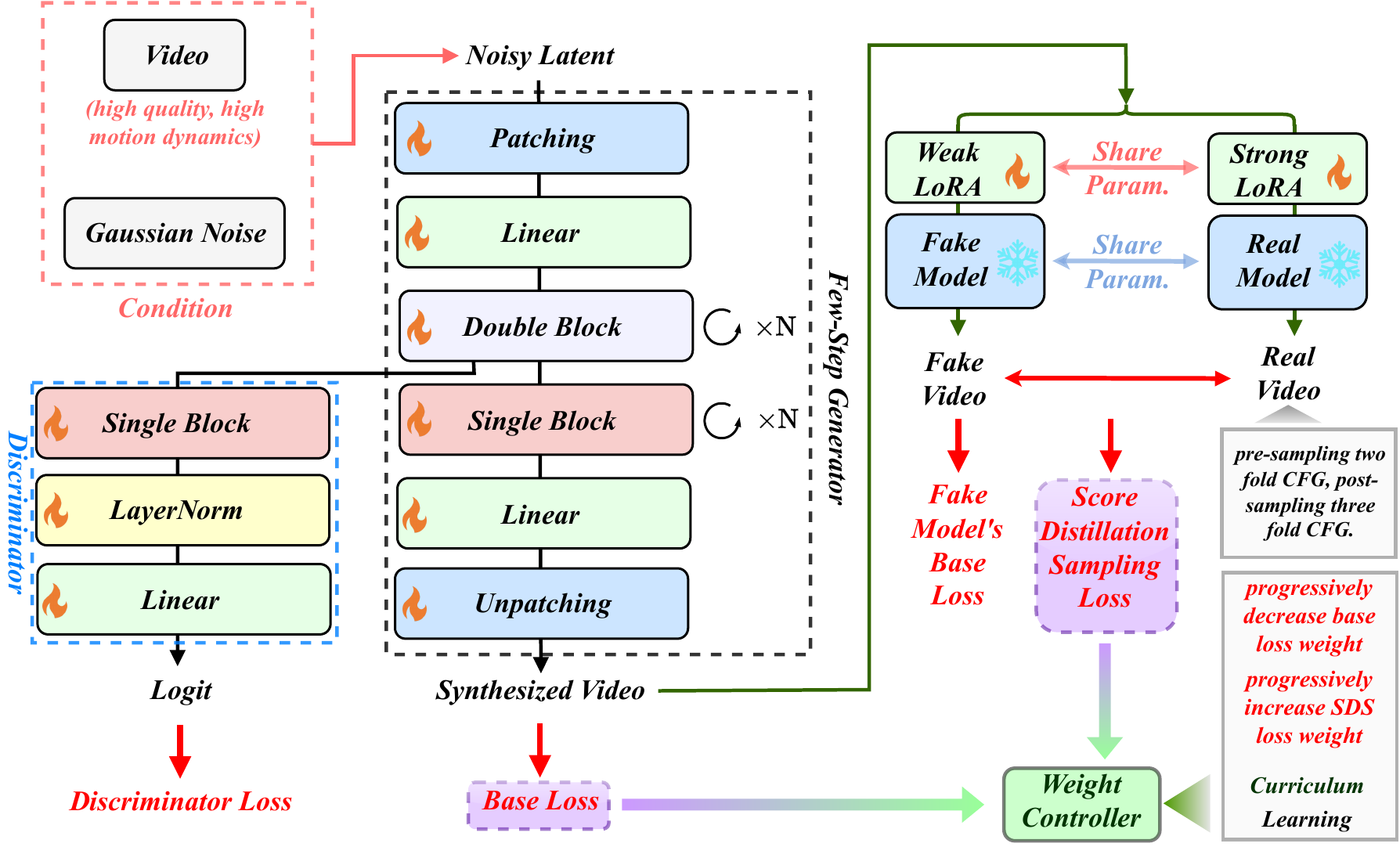}
    \caption{The overview of our modified DMD2. We employed a curriculum learning strategy to gradually reduce the weight of the base loss while progressively increasing the weight of the SDS loss, effectively avoiding abrupt shifts in learning objectives. Furthermore, we adopted a two-fold to three-fold CFG attenuation strategy in the calculation of the real data distribution, which significantly enhances the motion dynamics of the generated video.}
    \label{fig:dmd_and_cfg_distillation}
\end{figure}

The multi-step sampling process during diffusion model inference severely limits the model's inference speed, a problem that becomes even more pronounced in {\model_name}, a 13B diffusion model, due to the slow inference in each sampling step. As illustrated in Fig.~\ref{fig:dmd_and_cfg_distillation}, we achieve high-quality accelerated sampling through the collaborative distillation of DMD and CFG. Our baseline algorithm is DMD2. DMD2 is a cutting-edge algorithm inspired by score distillation sampling (SDS)~\citep{poole2023dreamfusion} designed to facilitate efficient distribution matching and achieve step distillation. Specifically, the training process of DMD2 involves the collaboration of three models: the one/four-step generator $G_\phi$, which undergoes parameter updates; the real {\model_name} $u_\theta^\textrm{real}$, responsible for estimating the real data distribution $p_\textrm{real}$; and the fake {\model_name} $u_\theta^\textrm{fake}$, which estimates the fake data distribution $p_\textrm{fake}$. Notably, all three models are initialized from the pre-trained {\model_name} to ensure consistency and efficiency throughout the process. Its distribution matching loss can be written as\begin{equation}
    \begin{split}
        &\nabla_\phi {L}_\textrm{SDS} \triangleq \mathbb{E}_{t} \nabla_\phi \mathcal{D}_\textrm{KL}\left(p_\textrm{fake}\Vert p_\textrm{real}\right) \approx - \mathbb{E}_t \int_\epsilon\Big([{z}_t -\sigma_t \\
        & {u}_\theta^\textrm{real}({z}_t,t,T_s,I_s,A) -  {z}_t+\sigma_t{u}_\theta^\textrm{fake}({z}_t,t,T_s,I_s,A)]\frac{\partial G_\phi(\epsilon)}{\partial \phi}d\epsilon\Big), \\
    \end{split}
    \label{eq:dmd_loss}
\end{equation}
where $z_t = \sigma_t z_1 + (1-\sigma_t)\hat{z}_0$ and $\hat{z}_0$ is synthesized from the few-step generator. $\sigma_t$ stands for the noise schedule. This equation transforms the original distribution matching on the score function (i.e., vanilla DMD2) into a novel distribution matching at $t=0$, aligning with the training paradigm of {\model_name}. Furthermore, DMD2 must update $u_\theta^\textrm{fake}$ in real time to ensure accurate estimation of $p_\textrm{fake}$:\begin{equation}
    L_\textrm{fake} =\mathbb{E}_{z_0,z_1,t \sim [0,1]} \bigg[ \Big\lVert {z_1} - \hat{z}_0-{u_\theta^\textrm{fake}} \big(z_t,t,T_s,I_s,A\big) \Big\lVert_2^2 \bigg].
        \label{eq:fake_loss}
\end{equation} 
The key difference between DMD2 and DMD lies in the addition of adversarial training, a technique employed to enhance and refine visual quality. In practice, we train our distillation model with 16 NVIDIA H100 GPUs (each with 80GB of memory). However, we found that even with ZeRO3 optimization, loading all three models simultaneously for standard DMD2 training was infeasible. To overcome this, we adopted an alternative approach: we leveraged the real {\model_name} $u_\theta^\textrm{real}$, augmented with low-rank adaptation (LoRA), to implement the fake {\model_name} $u_\theta^\textrm{fake}$. This adjustment enabled efficient minimization of Eq.~\ref{eq:fake_loss} using LoRA, offering a computationally viable solution to the resource constraints encountered.

Furthermore, directly applying $L_\textrm{SDS}$ does not guarantee the quality of the synthesized video or its motion dynamics. Moreover, directly incorporating a pre-trained video model for distribution matching can lead to training instability or result in an uncontrollable few-step generator producing low-quality videos. To address these challenges, we first gather real data with the highest possible quality and motion dynamics. Subsequently, we employ a curriculum learning strategy to gradually reduce the weight of $L_\textrm{base}$ while progressively increasing the weight of $L_\textrm{SDS}$, ensuring stable model training.

Finally, to achieve the collaborative distillation of CFG and steps, we leverage CFG during inference to estimate the real data distribution. This strategy was derived through empirical exploration, where a two-fold CFG is applied for timesteps between 0.75 and 1, while a three-fold CFG is utilized for timesteps ranging from 0 to 0.75. Specifically:
\begin{equation}
\begin{split}
    & \hat{u_\theta}^\text{two-fold} = (1+\omega_\text{audio}){u_\theta}(\mathbf{z}_t, t, T_s, I_s, A) - \omega_\text{audio}{u_\theta}(\mathbf{z}_t, t, \emptyset, \emptyset, \emptyset ), \\
    & \hat{u_\theta}^\text{three-fold} = (1+\omega_\text{audio}){u_\theta}(\mathbf{z}_t, t, T_s, I_s, A) - \omega_\text{audio}{u_\theta}(\mathbf{z}_t, t, T_s, I_s, \emptyset ) \\
    &\quad\quad\quad\quad+(1+\omega_\text{text}){u_\theta}(\mathbf{z}_t, t, T_s, I_s, \emptyset ) - \omega_\textrm{text}{u_\theta}(\mathbf{z}_t, t, \emptyset, \emptyset, \emptyset ), \\
\end{split}
\end{equation}
where $\omega_\textrm{audio}$ and $\omega_\textrm{text}$ represent the CFG scales specifically designed for the audio condition and text condition, respectively. In addition, our CFG scale is dynamic and will sample from a Gaussian distribution with mean values $\omega_\textrm{audio}$ and $\omega_\textrm{text}$. We found that this approach enhances the robustness of the few-step generator in synthesizing diverse characters and varying backgrounds.

\begin{figure*}
	\centering
	\includegraphics[width=0.99\linewidth]{Figure/exp_effect.pdf}
	\centering
	\caption{Qualitative experimental results of {\model_name}}
    \label{fig:exp_effect1}
\end{figure*}

\section{EXPERIMENT}
\subsection{Implementation Details}
\subsubsection{Datasets}
\label{subsub:datasets}
We train the model using internally collected data. 
All data were processed with a cropped resolution of 720${\times}$1280 and a video length of 129 frames, with the first frame serving as the portrait image during training.
To address performance degradation due to variations in video frame rates (fps), we train the {\model_name} model exclusively on videos resampled to a consistent 25 fps.
We use MediaPipe\cite{lugaresi2019mediapipe} for human face detection and tracking to extract face mask annotations, also filtering out videos with low-resolution faces, multiple faces or static content.
Videos with speakers occluded by subtitles or with audios of multiple speakers are also filtered using off-the-shelf tools.
Training textual instructions were derived from video descriptions extracted using LLAVA-34B~\cite{liu2024llava}, and rewritten by Gemini~\cite{team2023gemini}. 
A total of 1.85 million video clips are used for training.
To remove redundant computation during the training process, 3D VAE video features temporally aligned with Wav2Vec audio features, as well as reference image features, are also extracted during data preprocessing.
\subsubsection{Training}
We trained the base model of {\model_name} on a GPU server equipped with $200$ NVIDIA H100 GPUs, using the T2V model as the generative backbone. The training process for {\model_name} was conducted in two stages. In the first stage, we trained the model for 80,000 steps to generate videos guided by a static portrait image and textual features, excluding audio-related layers. In the second stage, we introduced audio cross-attention blocks—inserted at the end of each single and double block—and trained the model for an additional 30,000 steps. This enabled the model to animate the portrait image using both driving audio and textual prompts.

For the dual-stream phase, an audio cross-attention block was added after every double block, while in the single-stream phase, one was inserted after every two single blocks. We used the AdamW optimizer~\cite{diederik2014adam} with a learning rate of $10^{-5}$ for all modules across both stages. To emphasize facial local motion in the second stage, the adaptive loss weight $w$ was set to 10. Accelerated sampling was achieved through steps distillation and CFG distillation.

\subsubsection{Efficient Long Video Inference via Sliding Window Denoising}

For long-form video generation, we introduce a sliding window denoising approach that enables the synthesis of temporally coherent videos of arbitrary duration. Rather than generating the entire video at once—which would be computationally prohibitive—or generating independent segments that would result in temporal discontinuities, our method processes overlapping batches of frames while maintaining global consistency.

Given a long driving audio signal, our algorithm processes it through a multi-stage pipeline. First, the audio is preprocessed and padded to ensure clean transitions between segments. The temporal audio features are then resampled and aligned with the intended video frames, with careful handling of frame boundaries to prevent artifacts.

The core of our approach involves a progressive denoising diffusion process where a latent representation of the entire sequence is initialized and systematically denoised through a fixed number of diffusion steps. At each denoising timestep, we process overlapping batches of frames (33 frames per batch with a configurable overlap). For each batch, we predict noise conditioned on both the audio features and optional text prompts, applying classifier-free guidance separately for audio and text conditions. We implement an adaptive blending mechanism at the overlapping regions that ensures smooth transitions between consecutive batches.

This weighted blending strategy is particularly important, as it distributes the influence of each prediction according to a frame's position within the batch overlap. For frames in the overlapping regions, we compute:
\begin{equation*}
    w_{\text{orig}} = \frac{(w_{\text{overlap}} - 1) - i}{w_{\text{overlap}} - 2}, \quad w_{\text{new}} = 1 - w_{\text{orig}}
\end{equation*}

where $i$ represents the frame's index position in the overlap region and $w_{\text{overlap}}$ is the overlap width. Here, $w_{\text{orig}}$ is the weight applied to the previously computed frame and $w_{\text{new}}$ is the weight applied to the newly computed frame in the current batch.
Our approach reduces memory requirements while maintaining global temporal coherence, addressing the "infinite context" problem in long-form video generation. The model can maintain consistent identity and motion patterns throughout videos of arbitrary duration, without the need for explicit keyframe planning or segment-level supervision.

To further accelerate inference, we implemented Sequence Parallelism for Full-Attention computations using USP~\cite{fang2024USP}. This approach enables the efficient generation of a 1-minute animation (540${\times}$540 resolution, four-step sampling) within 60 seconds on 8 NVIDIA H100 GPUs.

\subsection{Evaluations and Comparisons}
\subsubsection{Qualitative analysis}
To more comprehensively assess the model's ability to animate free-style portraits and its applicability in various scenarios, we introduce a new benchmark, {\model_name}-Benchmark. 
Specifically, we collect 30 portraits from different domains, including various styles(e.g. anime, sculpture and realistic styles), which are generated using text-to-image models. 
These images featured adults, young people, teenagers, and infants, with backgrounds ranging from indoor settings (e.g., bedrooms, living rooms, classrooms) to outdoor environments (e.g., beaches, forests, streets). 

The {\model_name}-Benchmark also includes 20 audio clips representing different speaking scenarios such as singing, speech, and rapping, as well as 20 textual prompts that describe different emotions (e.g., happiness, anger, sadness) and actions (e.g., hand raising, head shaking) during speaking. Some textual prompts included descriptions of background changes specific to certain portrait images, such as leaves rustling in the wind or waves crashing on the beach. This benchmark, with its diverse styles and scenarios, will benefit the development of the community.

We evaluate {\model_name} on the {\model_name}-Benchmark. As shown in Figure ~\ref{fig:teaser}, {\model_name} demonstrates the superior perceptual quality and motion smoothness, even after model acceleration. {\model_name} exhibits superior domain generalization capabilities to style portraits, despite being trained exclusively on real portrait animations.

\begin{table}[h]
\begin{center}
\setlength{\tabcolsep}{1.5mm}
\renewcommand{\arraystretch}{1.1}
\begin{tabular}{l|p{6cm}}
\toprule
  IDs & Questions \\
\midrule
1 & Which video shows the best match between the character's lip movements and the audio? \\
\midrule
2 & Which video has the character that looks most like the person in the image? \\
\midrule
3 & Which video feels the smoothest and least choppy to you? \\
\midrule
4 & Which video has character movements that seem most realistic or similar to those in movies and animations? \\
\midrule
5 & Which video has scene changes that seem most realistic or similar to those in movies and animations? \\
\bottomrule
\end{tabular}
\end{center}
\caption{The user study comprises a list of five questions. Participants in the user study will answer these questions based on the videos synthesized by Hallo3, SadTalker, and {\model_name}, as well as the corresponding portrait images.}
\label{tab:user_study} 
\end{table}

We compared {\model_name} with prior portrait animation works, including state-of-the-art (SOTA) GAN-based methods (SadTalker~\cite{zhang2023sadtalker}) and recent diffusion-based approaches (Hallo3~\cite{cui2024hallo3}). To elucidate the distinctions between {\model_name} and the SOTA, we conducted a user study using the {\model_name}-Benchmark. Specifically, for each portrait image in the benchmark, we randomly selected a textual prompt and an audio clip to form a test case. We invited 30 participants from various regions worldwide to evaluate the quality of videos synthesized by {\model_name} and those generated by publicly available implementations of SOTA methods across all selected test cases. Each participant was asked to complete five questions, as illustrated in Tab.~\ref{tab:user_study}. They were instructed to select the synthesized videos that appeared more realistic and vivid. Among the 150 responses collected, \textbf{137 out of 150 (91.33\%)} participants confirmed that {\model_name} outperformed SadTalker~\cite{zhang2023sadtalker} and Hallo3~\cite{cui2024hallo3} in terms of video quality.

\subsubsection{Quantitative analysis}

\begin{table}[h]
\begin{center}
\setlength{\tabcolsep}{1.5mm}
\renewcommand{\arraystretch}{1.1}
\begin{tabular}{l|cc|cc}
\toprule
  Method & Sync-C (HDTF) $\uparrow$ & Sync-C (Internal) $\uparrow$ & Sync-D (HDTF) $\downarrow$ & Sync-D (Internal) $\downarrow$ \\
\midrule
SadTalker & 6.7526 & 4.4568 & 8.0753 & 9.9851 \\
Hallo3  & 6.7997 & 5.6112 & 8.6029 & 9.4386 \\
\midrule
\model_name & \textbf{7.2777} & \textbf{6.6943} & \textbf{7.9670} & \textbf{8.4012} \\
\bottomrule
\end{tabular}
\end{center}
\caption{Quantitative comparison of synchronization metrics on HDTF and internal data.}
\label{tab:quantitative_combined}
\end{table}

We evaluate the quality of generated portrait videos under different methods using widely adopted metrics: FID~\cite{heusel2017fid_metric}, FVD~\cite{unterthiner2019fvd_metric} and Sync~\cite{chung2017syncNet} (Sync-C and Sync-D). We randomly sample 100 video clips from HDTF~\cite{zhang2021hdtf} and internally collected data (which were not used in the model's training) as test videos. For each test video, we use the first frame as the static portrait image and generate the entire video, where the audio of the test video serves as the driving audio. The test video is used as the ground truth, and the textual prompt is extracted from the test video using LLAVA-34B~\cite{liu2024llava}, rewritten by Gemini~\cite{team2023gemini}, same to ~\ref{subsub:datasets}. As shown in Table ~\ref{tab:quantitative_combined}, {\model_name} consistently demonstrates superior image quality, motion accuracy, and lip synchronization accuracy compared to these baselines. Furthermore, with limited sampling steps and without CFG (after distillation), {\model_name} achieves a $20\times$ speedup, outperforming both SadTalker~\cite{zhang2023sadtalker} and Hallo3~\cite{cui2024hallo3}, while delivering competitive results with the base model.

\section{Conclusion}
We present {\model_name}, a novel DiT-based framework for portrait animation with precise audio-lip synchronization. It features (1) 3D Full-Attention with sliding window denoising for coherent, high-quality video synthesis; (2) a curriculum learning scheme integrating audio, text, and images for multi-modal control; (3) region-specific masks with adaptive loss for enhanced lip accuracy; and (4) distillation techniques for 20× faster inference. {\model_name} excels in sync and animation across diverse portraits, voices, and prompts.



\clearpage
{
\bibliographystyle{plain}
\bibliography{sample-bibliography}
}


\appendix

 

\end{document}